\pdfoutput=1

\documentclass[11pt]{article}

\usepackage[]{EMNLP2023}

\usepackage{times}
\usepackage{latexsym}

\usepackage{siunitx}
\usepackage{booktabs}
\usepackage[T1]{fontenc}

\usepackage[utf8]{inputenc}
\usepackage{graphicx}
\usepackage{url}
\usepackage{hyperref}
\usepackage{tcolorbox}
\usepackage{algorithm}
\usepackage[noend]{algpseudocode}

\usepackage{xcolor}
\usepackage{mdframed}
\usepackage{lipsum} 

\newmdenv[
  topline=True,
  bottomline=True,
  skipabove=\topsep,
  skipbelow=\topsep,
  leftmargin=0cm,
  rightmargin=0cm,
  innertopmargin=10pt,
  innerbottommargin=10pt,
  innerleftmargin=0pt,
  innerrightmargin=3pt,
  linewidth=2pt,
  linecolor=blue,
  backgroundcolor=gray!20
]{mybox}

\usepackage{microtype}

\usepackage{inconsolata}
\usepackage{multirow}
\usepackage{amsmath}
\usepackage{amssymb}
\usepackage{listings}
\title{\textit{`Quis custodiet ipsos custodes?'} Who will watch the watchmen? On Detecting AI-generated Peer Reviews }

\author{Sandeep Kumar$\dagger$, Mohit Sahu$\dagger$\thanks{$^*$Equal contribution.} , Vardhan Gacche$\dagger$\footnotemark[1] , Tirthankar Ghosal$\ddagger$,  Asif Ekbal$\S$ \\
  $\dagger$Indian Institute of Technology Patna, India \\
  $\ddagger$National Center for Computational Sciences, Oak Ridge National Laboratory, USA\\
  $\S$School of AI and Data Science,
IIT Jodhpur, India \\
$\dagger$\texttt{sandeep\_2121cs29@iitp.ac.in}, $\ddagger$\texttt{ghosalt@ornl.gov},
$\S$\texttt{asif@iitj.ac.in}} 

\begin{document}
\maketitle
\begin{abstract}
The integrity of the peer-review process is vital for maintaining scientific rigor and trust within the academic community. With the steady increase in the usage of large language models (LLMs) like ChatGPT in academic writing, there is a growing concern that AI-generated texts could compromise scientific publishing, including peer-reviews. Previous works have focused on generic AI-generated text detection or have presented an approach for estimating the fraction of peer-reviews that can be AI-generated. Our focus here is to solve a real-world problem by assisting the editor or chair in determining whether a review is written by ChatGPT or not. 
To address this, we introduce the Term Frequency (TF) model, which posits that AI often repeats tokens, and the Review Regeneration (RR) model, which is based on the idea that ChatGPT generates similar outputs upon re-prompting. We stress test these detectors against token attack and paraphrasing. Finally, we propose an effective defensive strategy to reduce the effect of paraphrasing on our models. Our findings suggest both our proposed methods perform better than the other AI text detectors. Our RR model is more robust, although our TF model performs better than the RR model without any attacks. We make our code, dataset, and model public\footnote{\url{https://github.com/sandeep82945/AI-Review-Detection}}\footnote{\url{https://www.iitp.ac.in/~ai-nlp-ml/resources.html}}.



\end{abstract}

\section{Introduction}
Large language models (LLMs), such as ChatGPT, PaLM \cite{chowdhery2023palm} and GPT-4 \cite{Achiam2023GPT4TR}, have significantly impacted both the industrial and academic sectors. The surge in Artificial Intelligence (AI)-generated content has permeated various domains, from journalism \cite{Gutiérrez-Caneda_Vázquez-Herrero_López-García_2023,journalmedia5020039} to academia \cite{BINNASHWAN2023102370,shi2023midmed}. However, their misuse
also introduces concerns—especially regarding fake news \cite{DBLP:conf/ijcnlp/ZhangG23, silva2024disinformation}, fake hotel
reviews \cite{ignat2024maide}, fake restaurant review \cite{gambetti2024aigen}. The exceptional human-like fluency and coherence of the generated content of these models pose a significant challenge, even for experts, in distinguishing if the text is written by humans or LLMs \cite{DBLP:journals/access/ShahidLSAHG22}. 


\begin{tcolorbox}[colback=gray!10!white, colframe=gray!60!black]
What if peer-reviews themselves are AI-generated? \textit{Who will guard the guards themselves?}
\end{tcolorbox}

A study \cite{DBLP:journals/corr/abs-2403-07183} conducted experiments on a few papers of AI conferences and found that between  6.5\% and 16.9\% of text submitted as peer-reviews to these conferences could have been substantially modified by LLMs. They estimated that the usage of ChatGPT in reviews increases significantly within three days of review deadlines. Reviewers who do not respond to ICLR/NeurIPS author rebuttals exhibit a higher estimated usage of ChatGPT. Additionally, an increase in ChatGPT usage is associated with low self-reported confidence in reviews. Once Springer retracted 107 cancer papers after they discovered that their peer-review process had been compromised by fake peer-reviewers \cite{Team_2022}.

In recent discussions surrounding the use of large language models (LLMs) in peer reviewing. According to  ACL policy\footnote{https://2023.aclweb.org/blog/review-acl23/\#faq-can-i-use-ai-writing-assistants-to-write-my-review}, if the focus is strictly on content, it seems reasonable to employ writing assistance tools for tasks such as paraphrasing reviews, particularly to support reviewers who are not native English speakers. However, it remains imperative that the reviewer thoroughly reads the paper and generates the review's content independently. Moreover, it is equally acceptable to use tools that assist with checking proofs or explaining concepts unfamiliar to the reviewer, provided these explanations are accurate and do not mislead the reviewer in interpreting the submission. This blend of automation and human oversight maintains the integrity of the review process while leveraging LLMs for specific enhancements. According to Elsevier policy\footnote{https://www.elsevier.com/en-in/about/policies-and-standards/the-use-of-generative-ai-and-ai-assisted-technologies-in-the-review-process}, reviewers should not upload their communications or any related material into an AI tool, even if it is just for the purpose of improving language and readability. They also emphasize that the critical thinking, original assessment, and nuanced evaluation required for a thorough review cannot be delegated to AI technologies, as these tools might produce incorrect, incomplete, or biased assessments. We believe reviewers should strictly adhere to the conference policy and guidelines regarding the use of AI tools in peer review, including for proofreading their reviews for refinement.

However, to the best of our knowledge, each venue agrees that the content of submissions and reviews is confidential. Therefore, they highly discourage the use of ChatGPT and similar non-privacy-friendly solutions for peer review. Additionally, they agree that AI-assisted technologies must not be used during the initial writing process of reviews. Consequently, our work aims to assist editors in identifying instances where reviewers may have bypassed this crucial step before using AI for refinement.



Previous works have focused on studying the effect of ChatGPT on AI conference peer-reviews. However, in this paper, our focus is to determine whether a review is written by ChatGPT or not. We do not assert that AI-generated peer-reviews inherently detract from the quality or integrity of the peer-review system. There can be debates whether AI-generated reviews can help peer-review system or not; we are not asserting that AI-generated peer-review is completely not useful. However, we believe if the review is AI-generated, the chair/meta-reviewer should be well aware. It is a breach of trust if the meta-reviewer believes that the review is human-written; nevertheless, it is not. Despite the potential benefits AI-gener, the chair/meta-reviewerated reviews may offer, it is crucial for editors to exercise discernment in their reliance on these reviews. This caution is warranted due to the intrinsic limitations of current language models, which can produce inaccurate, misleading \cite{DBLP:conf/emnlp/PanPCNKW23}, or entirely fabricated information—a phenomenon often referred to as hallucination \cite{ji-etal-2023-towards,rawte-etal-2023-troubling}.




In this paper, we propose two simple yet effective methods for detecting AI-generated peer reviews based on token frequency (TF method) and regeneration based approach (RR method). We also propose a token modification attack method and study its effect on various detectors. Paraphrasing attack is a very common way to evade text detection. So, we also study the effect of paraphrasing on various text detectors. Finally, we propose a technique to defend our regeneration-based technique against the paraphrasing attack. We found that both the TF model and the RR model perform better than other AI text detectors for this task. We also found that while the TF model performs better than the RR model under normal conditions, the RR model is more robust and is able to withstand adjective attacks and paraphrasing attacks (after the defense is applied).



\noindent We summarize our contributions as follows:-
\begin{itemize}
    \item We introduce a novel task to address the real-world problem of detecting AI-generated peer-reviews. We create a novel dataset of 1,480 papers from the ICLR and NeurIPS conferences for this task.
    \item We propose two techniques, namely the token frequency-based approach (TF) and the regeneration-based approach (RR), which perform better than the existing AI text detectors.
    \item We stress-test the detectors against token attacks and paraphrasing, and propose an effective defensive strategy to reduce evasion during paraphrasing attacks.
\end{itemize}

\section{Related Work}

\subsection{Zero-Shot Text Detection Detection}
Zero-shot text detection does not require training on specific data and directly identifies AI-generated text using the model that produced it \cite{DBLP:conf/icml/Mitchell0KMF23}. \cite{DBLP:journals/corr/abs-1908-09203} use average log probability of a text under the generative model for detection, whereas DetectGPT \citep{DBLP:conf/icml/Mitchell0KMF23} uses property of AI text to occupy negative curvature regions of model's log probability function. Fast-DetectGPT \citep{DBLP:journals/corr/abs-2310-05130} increases its efficiency by putting conditional probability curvature over raw probability. \citet{DBLP:conf/nips/TulchinskiiKKCN23} showed that the average intrinsic dimensionality of AI-generated texts is lower than that of human. The paper \citep{DBLP:conf/acl/GehrmannSR19} estimates the probability of individual tokens and detect AI-generated text by applying a threshold on probability.

\subsection{Training based Text Detection}
Some researchers have fine-tuned language models to recognize LLM-generated text. \citet{DBLP:journals/corr/abs-2301-07597} trained OpenAI text classifier on a collection on millions of text. GPT-Sentinel \citep{DBLP:journals/corr/abs-2305-07969} train RoBERTa \citep{DBLP:journals/corr/abs-1907-11692} and T5 \citep{DBLP:journals/jmlr/RaffelSRLNMZLL20} classifiers on OpenGPT-Text. LLM-Pat \citep{DBLP:journals/corr/abs-2305-12519} trained a neural network on the similarity between candidate texts and reconstructed sibling text generated by an intermediary LLM (parent). However, due to excessive reliance of this model on training data, many models show vulnerability to adversarial attacks \citep{DBLP:journals/corr/abs-2002-11768}.  

\subsection{LLM Watermarking}
The concept of watermarking AI-generated text, initially introduced by \cite{wiggers2022openai}, involves embedding an undetectable signal to attribute authorship to a particular text with a high level of confidence, which is similar to encryption and decryption.
In simple words, a watermark is a hidden pattern in text that is imperceptible to humans. It involves adding some kind of pattern which can be recognized by algorithms directly into the text and some techniques also involve integrating an machine learning model in the watermarking algorithm itself \citep{DBLP:conf/sp/AbdelnabiF21, DBLP:journals/corr/abs-2305-05773, DBLP:conf/acl/YooAJK23, DBLP:journals/ai/QiangZLZYW23}.

Watermarked text can be generated using a standard
language model without re-training \citep{DBLP:conf/icml/KirchenbauerGWK23}. It planted watermarks with large enough entropy, resulting in a change in the distribution of generated texts. \citet{DBLP:journals/corr/abs-2302-03162} proposed a method of injecting secret sinusoidal signals into decoding steps for each target token. 
However, \citet{DBLP:journals/corr/abs-2312-02382} addresses the issue that watermarking can compromise text generation quality, coherence, and depth of LLM responses. \citet{chakraborty2023counter} suggests that watermarked texts can be circumvented and paraphrasing
does not significantly disrupt watermark signals; thus, text watermarking is fragile and lacks reliability for real-life applications.

\subsection{Statistical Estimation Approach}
There have been inquiries into the theoretical feasibility of achieving precise detection on an individual level \citep{DBLP:journals/corr/abs-2306-15666,DBLP:journals/corr/abs-2303-11156,DBLP:journals/corr/abs-2304-04736}. \citep{DBLP:journals/corr/abs-2403-07183} presented an approach for estimating the fraction of text in a large corpus using a maximum likelihood estimation of probability distribution without performing inference on an individual level thus making it computationally efficient. They conducted experiments on papers from a few AI conferences to determine the fraction of peer-reviews that could have been substantially modified by LLMs.

\subsection{AI-generated Research Paper Detection}


The DagPap22 Shared Task \cite{kashnitsky2022overview} aimed to detect automatically generated scientific papers.
The dataset includes both human-written and likely AI-generated texts, with around 69\% being "fake," some generated by {\href{https://pdos.csail.mit.edu/archive/scigen/}{SCIgen.}}
The winning team \cite{rosati2022synscipass} utilized a DeBERTa v3 model that was fine-tuned on their dataset (almost all teams managed to surpass the baseline models, Tf-IDF and logistic regression). It was also concluded that machine-generated text detectors should not be used in production because they perform poorly with distribution shifts, and their effectiveness on realistic full-text scientific manuscripts remains untested.

There have been many efforts to improve the peer review process \cite{9651781,ghosal2022peer,kumar2022deepaspeer,li2020multi,kumar2023deepmetagen,kumar2023apcs,kang18naacl,kumar2023mup,darcy-etal-2024-aries}; however, AI-generated review text poses a significant challenge. To the best of our knowledge, we are the first to propose techniques specifically for the detection of AI-generated peer reviews.

\section{Dataset}
We collected a total of 1,480 papers from OpenReview Platform \footnote{\url{https://openreview.net/}}. The first version of ChatGPT was released by OpenAI on November 30, 2022. Therefore, we choose papers from 2022, ensuring there was almost no chance that any of the collected reviews were already generated by ChatGPT.

\begin{figure}[t]
\centering
\includegraphics[width=0.30\textwidth]{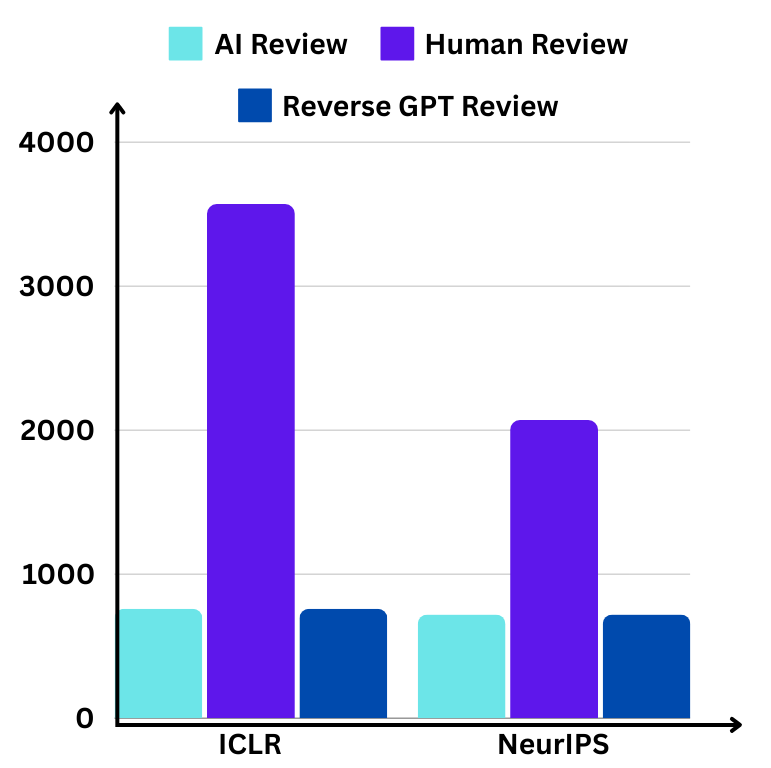}
    \caption{Dataset Statistics. Here, x axis: Different Venue ; y axis: Number of reviews.}
    \label{fig:Dataset}
\end{figure}
Figure \ref{fig:Dataset} shows the overall statistics of AI-generated reviews and golden reviews for both ICLR and NeurIPS reviews. We discuss the creation of the dataset in more details in the Appendix Section \ref{Appendix: dataset}. We split the dataset into 70\%, 15\%, and 15\% for training validation and test set respectively.

\section{Methodology}
In this section, we present our two approaches
to detect AI-written peer-reviews based on token frequency (Section \ref{Sec: method1}) and review regeneration (Section \ref{Sec: method2}). Then, we propose a possible attack (Token Manipulation Attack) on the AI text detectors to see how various models react to it in Section \ref{Sec: AI word Attack}. Additionally, since paraphrasing is a common method used to circumvent AI text detection, we introduce a countermeasure as described in Section \ref{Section: Paraphrasing Defence}, designed to protect our proposed Review Regeneration method against such attacks.

\subsection{Token Frequency based Approach} \label{Sec: method1}
Inspired by \cite{Liang2024MonitoringAC}, we propose a method that utilizes the frequency of tokens within review texts. This approach is premised on the hypothesis that different types of reviews (human-generated vs. AI-generated) exhibit distinct patterns in the usage of certain parts of speech, such as adjectives, nouns, and adverbs.

Let \( H \) denote the human corpus, consisting of all human-generated reviews, and \( A \) represent the AI corpus, comprising of all AI-generated reviews. Define \( x \) as an individual review, and \( t \) as a token. This token \(t\) can be adjective or noun or adverb. To identify if the token is adjective or noun or adverb, we have used the PoS-tagger of Natural Language Tool Kit (NLTK) module \footnote{\url{https://www.nltk.org/book/ch05.html}}. 


We define \( p^{A}(t) \) and \( p^{H}(t) \) as the probabilities of token \( t \) appearing in the AI and human corpora, respectively. These are estimated as follows:

\[
p^{A}(t) = \frac{\text{Count of reviews with } t \text{ in } A}{\text{Total \# of reviews in } A}
\]

\[
p^{H}(t) = \frac{\text{Count of reviews with } t \text{ in } H}{\text{Total \# of reviews in } H}
\]


Now, for each review \textit{x}, we calculate \textit{\(P^{A}(x)\) } and \textit{\(P^{H}(x)\)}, which represent the probability of \textit{x} belonging to the AI corpus and the human corpus, respectively. These probabilities can be calculated by summing up the probabilities of all tokens that are coming in review \textit{x}:-  
    \[P^{A}(x) = p^{A}(t_1) + p^{A}(t_2) + ... = \sum_{i=1}^{i=n_a} p^{A}(i) \]
    
    \[P^{H}(x) = p^{H}(t_1) + p^{H}(t_2) + ... = \sum_{i=1}^{i=n_h} p^{H}(i)\]
Here, \(t_1, t_2, ...\) refer to the tokens occurring in review \textit{x}. Also, \(n_a\) and \(n_h\) refer to the number of AI and Human corpus reviews, respectively.

If review \textit{x} contains tokens with higher probabilities in the AI corpus, then \textit{\(P^{A}(x)\)} will be greater, increasing the likelihood that \textit{x} is AI-generated. Conversely, if \textit{x} contains tokens with higher probabilities in the human corpus, then \textit{\(P^{H}(x)\)} will be greater, suggesting that the review is more likely to be human-written.

To classify each review \(x_i\), we calculate \(p^{A}(i)\) and \(p^{H}(i)\) for each review in our dataset. These serve as input features for training a neural network. The neural network is trained to distinguish between AI-generated and human-generated reviews based on these input features. By learning from the patterns and distributions of these probabilities, the neural network can accurately detect AI-generated reviews.

\subsection{Regeneration based Approach} \label{Sec: method2}

\begin{figure}[h]
\centering
\includegraphics[width=0.49\textwidth]{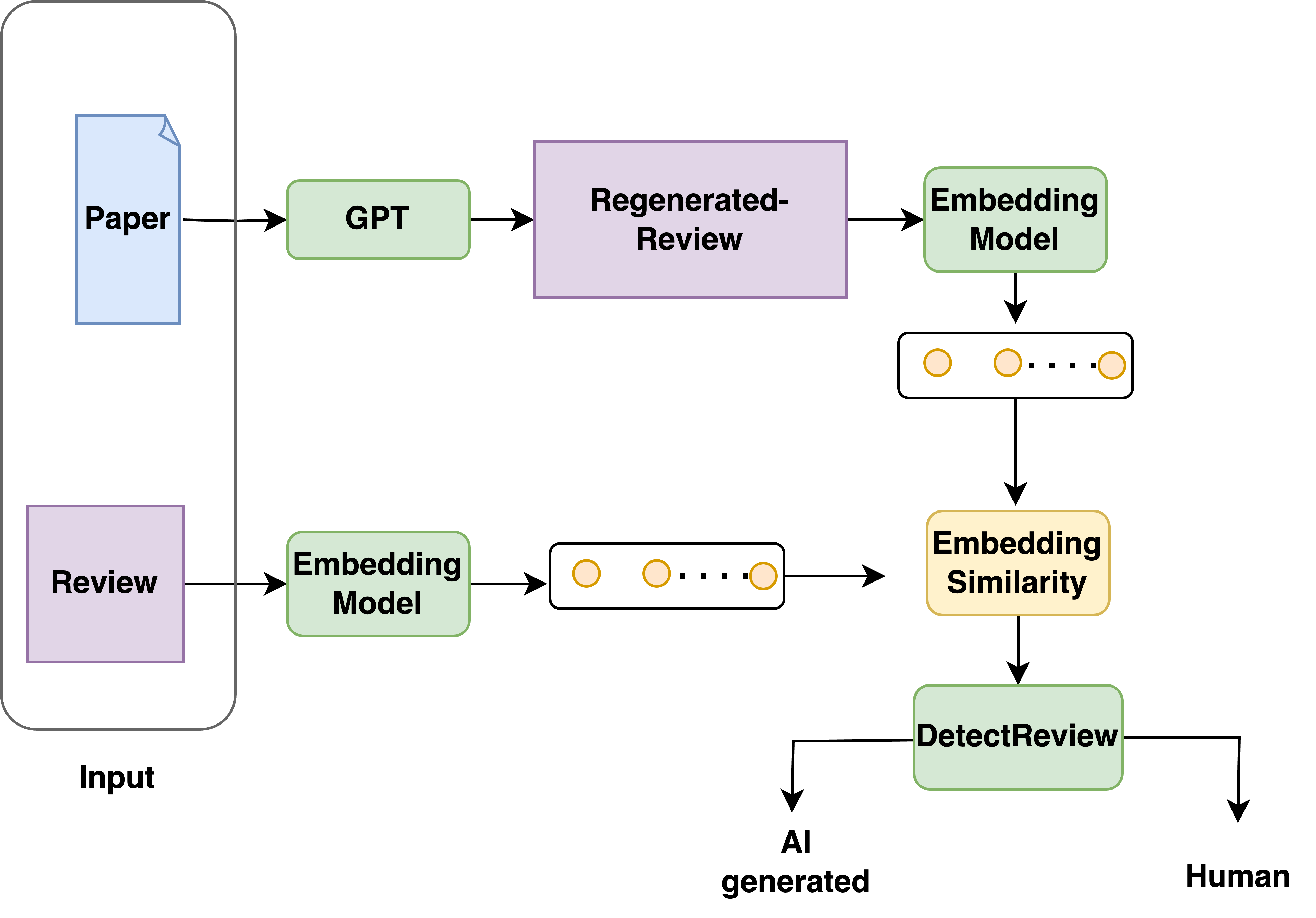}
    \caption{Architectural diagram of Regeneration based Approach.}
    \label{fig: method2}
\end{figure}


Figure \ref{fig: method2} shows the overall architectural diagram of our proposed regeneration-based approach. The input to the framework is the paper and its review which we aim to determine whether they are written by AI or Human.

The idea behind this approach is that if a similar prompt is given repeatedly to a large language model (LLM), the LLM is likely to generate reviews or responses that exhibit a consistent style, tone, and content, as outlined in the provided context. This consistency occurs because a large language model generally applies the patterns it has learned during training to the new content it generates based on the given prompt. The study in \cite{DBLP:journals/corr/abs-2308-02575} found that GPT-4 demonstrated high inter-rater reliability, with ICC scores ranging from 0.94 to 0.99, in rating responses across multiple iterations and time periods (both short-term and long-term). This indicates consistent performance when given the same prompt. Furthermore, the results showed that different types of feedbacks (content or style) did not affect the consistency of GPT-4’s ratings, further supporting the model’s ability to maintain a consistent approach based on the prompt.

\subsubsection{Review Regeneration and Embedding Creation}
We employ GPT to regenerate a review \( R^{reg} \) using the prompt  \( P^{reg} \).
We create two distinct embeddings $E_{R}$ for \( R^{reg} \) and $E_{F}$ for $R$ (review which we have to determine if the review is AI-generated or not). The idea is that if the review \(R\) is generated by an AI, we hypothesize that its embedding \(E_{F}\) will exhibit a closer similarity to \(E_{R}\), the embedding of a known AI-generated review \(R^{reg}\).

Then, we quantify the similarity between the embeddings using the cosine similarity metric, as outlined below:
\[
\text{CosineSimilarity}(E_{R}, E_{F}) = \frac{E_{R} \cdot E_{F}}{\|E_{R}\| \|E_{F}\|}
\]

Here, $\cdot$ represents the dot product, and $\|R\|$ and $\|F\|$ represent the Euclidean norms of the embeddings. This formula calculates the cosine of the angle between the two embeddings \(E_{R}\) and \(E_{F}\), providing a measure of similarity where values closer to 1 indicate higher similarity and thus a greater likelihood that both reviews are AI-generated.

\subsubsection{Training}





Next, we utilize the computed similarity score as input to train a neural network aimed at detecting AI-generated reviews. The training process involves optimizing the network's parameters via backpropagation. This optimization is directed by the cross-entropy loss function.




\subsection{Token Attack} \label{Sec: AI word Attack}

\begin{figure}[ht]
\centering
\includegraphics[width=0.49\textwidth]{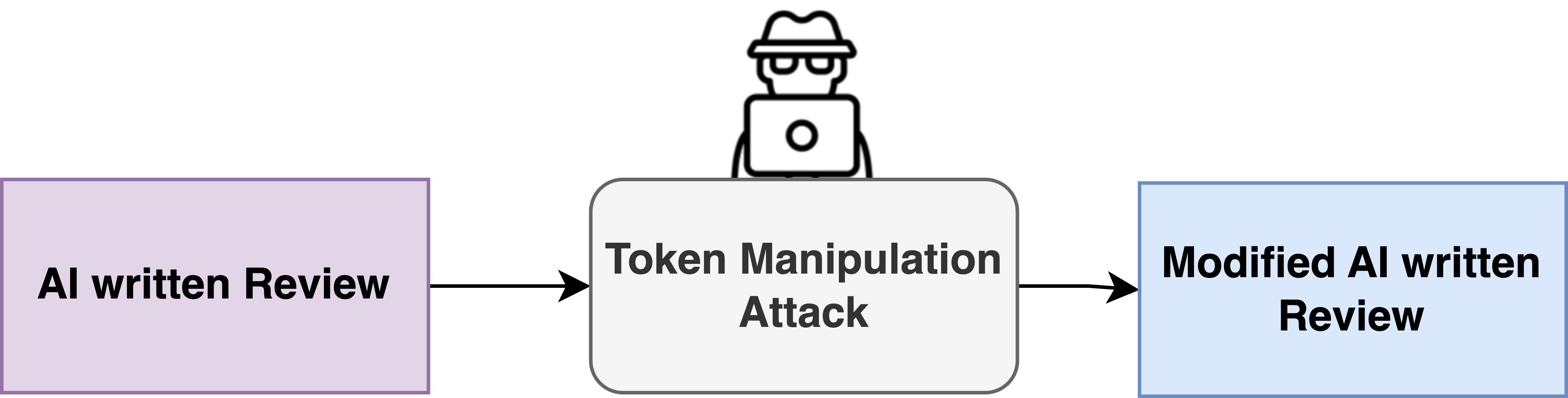}
    \caption{AI text undetectability attack.}
    \label{fig: attack}
\end{figure}

\begin{figure}[ht]
\centering
\includegraphics[width=0.49\textwidth]{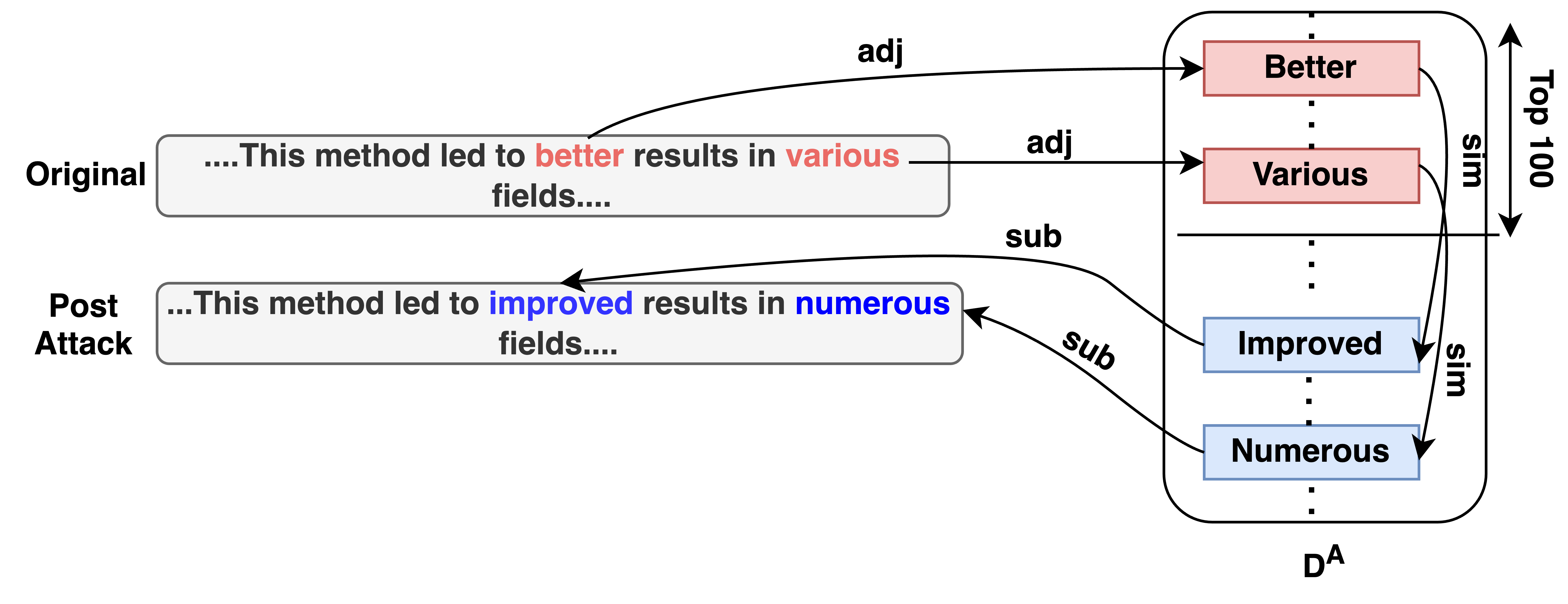}
    \caption{An example of adjective token attack. Here, sub: substitution, adj: Adjective, sim: similar token , $D^{A}$ : AI word dictionary (sorted high-top to bottom-low).}
    \label{fig: word_attack}
\end{figure}

We propose an attack method to reduce the probability of reviews being classified as AI-generated described in \hyperref[alg:word_attack]{Algorithm-1} where we target the most frequent tokens in AI-generated reviews and replace them with their synonyms, which are less frequent in the AI-generated content. 

Here, we focus exclusively on adjectives, referring to this approach as the "adjective attack." We chose adjectives because substituting nouns and adverbs with their synonyms often leads to nonsensical statements or drastically alters the meaning of the review. We discuss this in detail in Appendix \ref{Appendix: token_attack}. 


In the adjective attack, we substitute the top 100 highest probability adjective tokens (e.g., "novel," "comprehensive") with their synonyms. 


To obtain synonyms for the selected tokens, we utilize the NLTK WordNet database\footnote{\url{https://www.nltk.org/api/nltk.corpus.reader.wordnet}}. To preserve the original meaning of tokens as much as possible, we ensure that any synonym used to replace a token is also present in the AI corpus. If a suitable synonym is not found in the corpus, we do not replace the token.


\begin{algorithm}
\caption{Token Attack}\label{alg:word_attack}
\begin{algorithmic}[1]

\State Identify top 100 high-probability tokens: $w_1, w_2, \ldots, w_{100}$.
\State Retrieve synonyms for each token: $sw_1, sw_2, \ldots, sw_{100}$.
\State Perform PoS tagging for each review
\State Replace each tagged token with its synonym if it matches with one of the top 100 tokens.
\end{algorithmic}
\end{algorithm}

In order to determine which tokens from the review should be replaced with their synonyms, we performed PoS tagging on the review. For example, if we are conducting an adjective attack, we replace only the adjective tokens in the review with their synonyms.



We also illustrate this with an example of an adjective attack, as shown in Figure \ref{fig: word_attack}. In this example, the adjective tokens `better' and `various' from a review are among the top 100 AI token list. We replace them with their synonyms, `improved' and `numerous,' respectively.

\subsection{Paraphrasing Defence} \label{Section: Paraphrasing Defence}

Paraphrasing tools are effective in evading detection \cite{sadasivan2023can,krishna2024paraphrasing}. Given the fluency and coherence of paraphrased content, it is hard to tell if the text is written by a human or AI even for experts. To increase the robustness of Regeneration based text detector to paraphrase attacks, we introduce a simple defense that employs a targeted synonym replacement strategy. The core idea behind this approach is that when an AI-generated review is processed by a paraphraser, one of the major modifications it makes is substituting the original words with similar ones. We propose a technique to revert the paraphrased reviews back to a state that closely resembles their original AI-generated form by utilizing the regenerated review (as they would be close to the original AI-generated review).




\begin{algorithm}
\caption{Paraphrasing Defence}
\begin{algorithmic}[1]
        \State Identify tokens in the review and regenerated reviews
        \For{each token in the review}
            \State Get synonyms of the token
            \For{each synonym in synonyms}
                \If{synonym is in regenerated reviews}
                    \State Replace the token with synonym
                    \State Break
                \Else
                    \State Do not replace the token
                \EndIf
            \EndFor
        \EndFor
\end{algorithmic}
\label{alg:paraphrase_attack}
\end{algorithm}

As discussed in \hyperref[alg:paraphrase_attack]{Algorithm-2},
first, we identify all the tokens within a review and their corresponding regenerated reviews using the PoS tagging\footnote{We used tagger of the NLTK model. As we also discussed in Section \ref{Sec: AI word Attack}}. Here token can be any word in a review which are adjective, noun, or adverb. For each token in a review, we obtain a list of synonyms from the NLTK WordNet database. Then, for each synonym in that list, we check whether it is present in the corresponding regenerated review or not. If it is, we replace the original token with its synonym.

\begin{figure}[ht]
\centering
\includegraphics[width=0.49\textwidth]{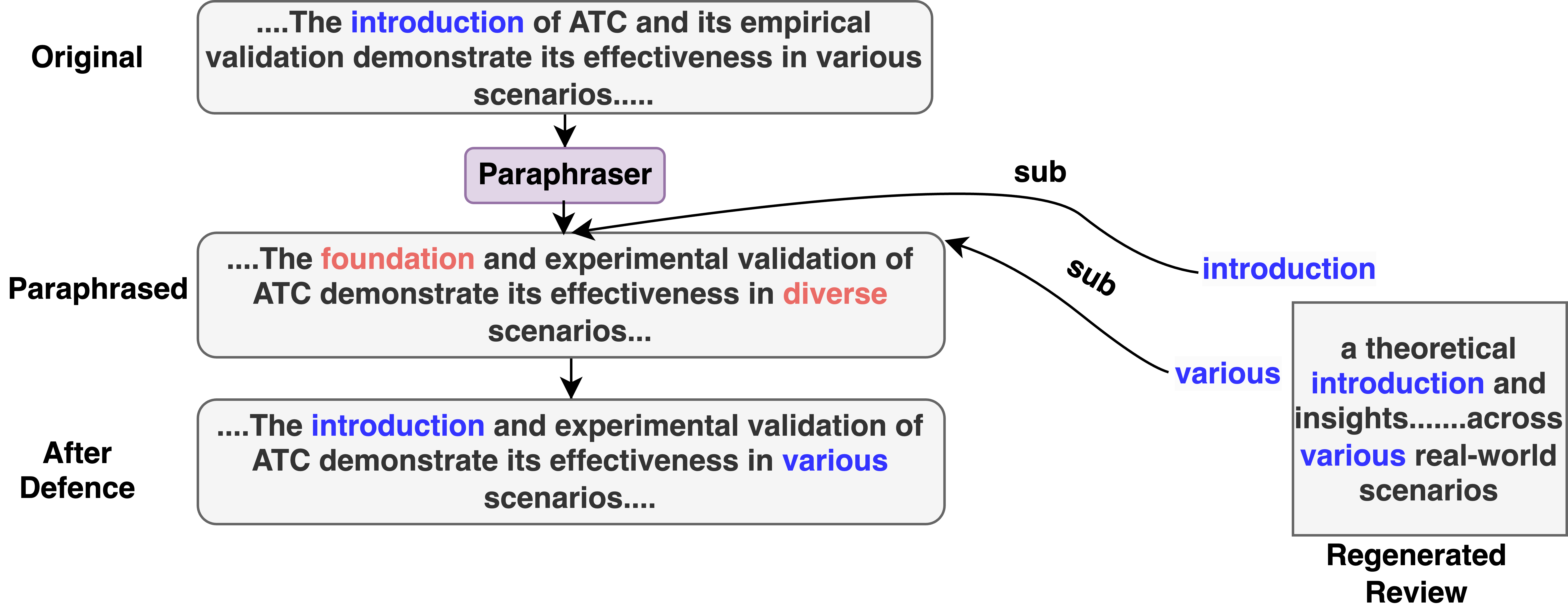}
    \caption{An example of paraphrasing defence; Here,sub: substitution.}
    \label{fig: word_defence}
\end{figure}

We also illustrate this by an example in Figure \ref{fig: word_defence}. The paraphraser has changed the structure of the sentence and also replaced some of the words like `introduction' with `foundation', `empirical' with `experimental,' and `various' with `diverse'. Now, after applying the defence algorithm the words `foundation' and `diverse' gets reverted back to `introduction' and `various', thus making it more identical to its original sentence. We called a review converted by using this algorithm as 'modified review'.

\textbf{Training:}  In a real-world scenario, whether a review has been paraphrased or not will be unknown, and detecting this becomes a task in itself. However, the aim of this paper is to propose a model that is robust to any kind of text, whether paraphrased or not. Therefore, we retrained both models. The modified training set consists of the original training set after being processed by the defense algorithm. Similarly, the modified paraphrased set consists of the paraphrased reviews from the original training set, which have been modified using the defense algorithm. For testing or validation, it will be unclear whether a review is paraphrased by AI or simply AI-written. Therefore, we combined both the testing set and the paraphrased set. Both will be modified by the defense algorithm before undergoing validation or testing\footnote{As a result, the size of the training set will increase threefold, and the testing and validation sets will double}.




\begin{table*}[ht]
\centering
\label{tab:my-table}
\begin{tabular}{|l|llllllll|}
\hline
\multirow{2}{*}{\textbf{Model}}
                                & \multicolumn{2}{c|}{\textbf{Precision}}                                 & \multicolumn{2}{c|}{\textbf{Recall}}                                    & \multicolumn{2}{c|}{\textbf{F1 - Score}}                                & \multicolumn{2}{c|}{\textbf{Accuracy}}             \\ \cline{2-9} 
                                & \multicolumn{1}{l|}{\textbf{ICLR}} & \multicolumn{1}{l|}{\textbf{NeurIPS}} & \multicolumn{1}{l|}{\textbf{ICLR}} & \multicolumn{1}{l|}{\textbf{NeurIPS}} & \multicolumn{1}{l|}{\textbf{ICLR}} & \multicolumn{1}{l|}{\textbf{NeurIPS}}  &\multicolumn{1}{l|}{\textbf{ICLR}} & \textbf{NeurIPS} \\ \hline
\textbf{RADAR}                  & \multicolumn{1}{l|}{66.48}        & \multicolumn{1}{l|}{66.97}        & \multicolumn{1}{l|}{75.13}        & \multicolumn{1}{l|}{81.11}        & \multicolumn{1}{l|}{70.54}        & \multicolumn{1}{l|}{73.37}        & \multicolumn{1}{l|}{66.12}        & 69.01        \\ \hline
\textbf{LLMDET}                 & \multicolumn{1}{l|}{54.69}        & \multicolumn{1}{l|}{53.24}        & \multicolumn{1}{l|}{98.42}        & \multicolumn{1}{l|}{98.06}        & \multicolumn{1}{l|}{70.30}        & \multicolumn{1}{l|}{69.01}        & \multicolumn{1}{l|}{55.11}        & 53.65        \\ \hline
\textbf{DEEP-FAKE}              & \multicolumn{1}{l|}{93.98}        & \multicolumn{1}{l|}{93.64}        & \multicolumn{1}{l|}{92.50}        & \multicolumn{1}{l|}{91.94}        & \multicolumn{1}{c|}{93.24}        & \multicolumn{1}{l|}{92.78}        & \multicolumn{1}{l|}{89.45}        & 88.89        \\ \hline
\textbf{FAST DETECT}            & \multicolumn{1}{l|}{95.96}        & \multicolumn{1}{l|}{94.87}        & \multicolumn{1}{l|}{81.32}        & \multicolumn{1}{l|}{66.81}        & \multicolumn{1}{l|}{88.03}        & \multicolumn{1}{l|}{78.40}        & \multicolumn{1}{l|}{88.07}        & 80.63        \\ \hline

\textbf{Our TF Model}                                       & \multicolumn{1}{l|}{\textbf{99.99}}             & \multicolumn{1}{l|}{\textbf{99.99}}             & \multicolumn{1}{l|}{\textbf{99.80}}         & \multicolumn{1}{l|}{\textbf{99.30}}         & \multicolumn{1}{l|}{\textbf{99.89}}         & \multicolumn{1}{l|}{\textbf{99.65}}        & \multicolumn{1}{l|}{\textbf{99.92}}        & \textbf{99.82}                                                     \\ \hline
\multicolumn{1}{|l|}{\textbf{Our RR Model}}   & \multicolumn{1}{l|}{99.32}                                & \multicolumn{1}{l|}{93.75}                                & \multicolumn{1}{l|}{94.38}                                & \multicolumn{1}{l|}{93.10}                                 & \multicolumn{1}{l|}{96.79}                                & \multicolumn{1}{l|}{93.43}                                & \multicolumn{1}{l|}{98.67}                                & 97.24 \\ \hline
\end{tabular}
\caption{Comparison results of the proposed Review Regeneration technique and Token Frequency technique. Here, the AI-generated reviews and regenerated reviews are generated by GPT-4; RR: Review Regeneration, TF: Token Frequency.}
\label{Tab: result-1}
\end{table*}

\begin{table*}[ht]
\centering
\begin{tabular}{|l|llllllll|}
\hline

                                 & \multicolumn{2}{c|}{\textbf{Precision}}                                                         & \multicolumn{2}{c|}{\textbf{Recall}}                                                            & \multicolumn{2}{c|}{\textbf{F1-Score}}                                                          & \multicolumn{2}{c|}{\textbf{Accuracy}}                                                          \\ \cline{2-9} 

\multirow{-3}{*}{\textbf{Model}} & \multicolumn{1}{c|}{\textbf{ICLR}} & \multicolumn{1}{c|}{\textbf{NeurIPS}} & \multicolumn{1}{c|}{\textbf{ICLR}} & \multicolumn{1}{c|}{\textbf{NeurIPS}} & \multicolumn{1}{c|}{\textbf{ICLR}} & \multicolumn{1}{c|}{\textbf{NeurIPS}} & \multicolumn{1}{c|}{\textbf{ICLR}} & \multicolumn{1}{c|}{\textbf{NeurIPS}} \\ \hline

\textbf{RADAR}                                           & \multicolumn{1}{l|}{14.58}        & \multicolumn{1}{l|}{15.13}        & \multicolumn{1}{l|}{40.38}        & \multicolumn{1}{l|}{51.11}        & \multicolumn{1}{l|}{21.43}        & \multicolumn{1}{l|}{23.35}        & \multicolumn{1}{l|}{47.97}        & 48.99                                                     \\ \hline

\textbf{LLMDET}                                          & \multicolumn{1}{l|}{50.17}        & \multicolumn{1}{l|}{52.53}        & \multicolumn{1}{l|}{\textbf{95.39}}        & \multicolumn{1}{l|}{\textbf{93.75}}        & \multicolumn{1}{l|}{65.76}        & \multicolumn{1}{l|}{67.33}        & \multicolumn{1}{l|}{50.33}        & 52.88                                                     \\ \hline
\textbf{DEEP-FAKE}                                       & \multicolumn{1}{l|}{68.42}        & \multicolumn{1}{l|}{47.37}                                & \multicolumn{1}{l|}{17.11}        & \multicolumn{1}{l|}{93.06}                                & \multicolumn{1}{l|}{27.37}        & \multicolumn{1}{l|}{11.04}                                & \multicolumn{1}{l|}{54.61}        & 49.65                                                     \\ \hline
\textbf{FAST DETECT}                                     & \multicolumn{1}{l|}{71.43}                                & \multicolumn{1}{l|}{20.00}                                & \multicolumn{1}{l|}{03.47}                                & \multicolumn{1}{l|}{00.69}                                & \multicolumn{1}{l|}{06.62}                                & \multicolumn{1}{l|}{01.34}                                & \multicolumn{1}{l|}{51.04}                                & 48.96                                                     \\ \hline
\multicolumn{1}{|l|}{\textbf{Our TF Model}}               & \multicolumn{1}{l|}{\textbf{99.99}}                                & \multicolumn{1}{l|}{\textbf{99.99}}                                & \multicolumn{1}{l|}{11.18}                                     & \multicolumn{1}{l|}{05.56}                                     & \multicolumn{1}{l|}{20.12}                                & \multicolumn{1}{l|}{10.53}                                & \multicolumn{1}{l|}{81.45}                                & 79.35                                                    \\ \hline
\multicolumn{1}{|l|}{\textbf{Our RR Model}}                 & \multicolumn{1}{l|}{81.67}                                & \multicolumn{1}{l|}{80.87}                                & \multicolumn{1}{l|}{64.47}                                & \multicolumn{1}{l|}{64.58}                                & \multicolumn{1}{l|}{\textbf{72.06}}                                & \multicolumn{1}{l|}{\textbf{71.81}}                                & \multicolumn{1}{l|}{\textbf{89.78}}                                & \textbf{89.23}                                                     \\ \hline
\end{tabular}
\caption{Comparison results after Token Attack (Adjective).}
\label{Tab: result_att_adj}
\end{table*}

\section{Experiments}
\subsection{Experimental Settings}
We implemented our system using PyTorch \cite{DBLP:conf/nips/PaszkeGMLBCKLGA19}. The dataset was randomly split into three parts: 80\% for training, 10\% for validation, and 10\% for testing.

For the TF model and RR model, we conducted experiments with different network configurations during the validation phase. Through these experiments, we determined that a batch size of 32 and a dropout rate of 0.1 for every layer yielded optimal performance. The activation function ReLU was used in our model. We trained the model for 20 epochs, employing a learning rate of 1e-3 for TF model and 0.01 for RR model and cross-entropy as the loss function. To prevent overfitting, we used the Adam optimizer with a weight decay of 1e-3. We  trained all the models on an NVIDIA A100 40GB GPU. We used the text-embedding-ada-002\footnote{\url{https://platform.openai.com/docs/guides/embeddings}} pretrained model from OpenAI for creating embeddings of the reviewer's review and the regenerated review.

\subsection{Baselines for Comparison}


 \textbf{RADAR \cite{DBLP:conf/nips/HuCH23}} (Robust AI text Detection via Adversarial Learning) draws inspiration from adversarial machine learning techniques.    
    \textbf{LLMDet \cite{wu2023llmdet}} (A Third Party Large Language Models Generated Text Detection Tool)
     is a text detection tool that can identify the source from which the text was generated, such as Human, LLaMA, OPT, or others. 
    \textbf{DEEP-FAKE \cite{li2023deepfake}} Text Detection considered 10 datasets covering a
wide range of writing tasks (e.g., story generation,
news writing and scientific writing) from diverse
sources (e.g., Reddit posts and BBC news), and applied
27 LLMs (e.g., OpenAI, LLaMA, and EleutherAI)
for construction of deepfake texts.
    \textbf{Fast-Detect GPT \cite{bao2023fast}} uses a conditional probability function and it invokes the sampling GPT once to generate all samples and calls the scoring GPT once to evaluate all the samples. We discuss them in details in Section \ref{Appendix: baseline}.

\begin{table*}[ht]
\centering
\begin{tabular}{|l|llllllll|}
\hline

                                 & \multicolumn{2}{c|}{\textbf{Precision}}                                                         & \multicolumn{2}{c|}{\textbf{Recall}}                                                            & \multicolumn{2}{c|}{\textbf{F1-Score}}                                                          & \multicolumn{2}{c|}{\textbf{Accuracy}}                                                          \\ \cline{2-9} 

\multirow{-3}{*}{\textbf{Model}} & \multicolumn{1}{c|}{\textbf{ICLR}} & \multicolumn{1}{c|}{\textbf{NIPS}} & \multicolumn{1}{c|}{\textbf{ICLR}} & \multicolumn{1}{c|}{\textbf{NIPS}} & \multicolumn{1}{c|}{\textbf{ICLR}} & \multicolumn{1}{c|}{\textbf{NIPS}} & \multicolumn{1}{c|}{\textbf{ICLR}} & \multicolumn{1}{c|}{\textbf{NIPS}}                                                   \\ \hline
\textbf{RADAR}                                           & \multicolumn{1}{l|}{88.82}        & \multicolumn{1}{l|}{95.83}        & \multicolumn{1}{l|}{51.92}        & \multicolumn{1}{l|}{53.08}        & \multicolumn{1}{l|}{65.53}        & \multicolumn{1}{l|}{68.32}        & \multicolumn{1}{l|}{53.29}        & 55.56                                                         \\ \hline

\textbf{LLMDET}                                          & \multicolumn{1}{l|}{\textbf{98.68}}        & \multicolumn{1}{l|}{\textbf{99.31}}        & \multicolumn{1}{l|}{49.83}        & \multicolumn{1}{l|}{50.00}        & \multicolumn{1}{l|}{66.23}        & \multicolumn{1}{l|}{66.51}        & \multicolumn{1}{l|}{49.67}        & 50.00                                                    \\ \hline
\textbf{DEEP-FAKE}                                       & \multicolumn{1}{l|}{83.55}        & \multicolumn{1}{l|}{78.47}                                & \multicolumn{1}{l|}{70.17}        & \multicolumn{1}{l|}{60.75}                                & \multicolumn{1}{l|}{76.28}        & \multicolumn{1}{l|}{68.48}                                & \multicolumn{1}{l|}{74.01}        & 63.89                                                     \\ \hline
\textbf{FAST DETECT}                                     & \multicolumn{1}{l|}{59.35}                                & \multicolumn{1}{l|}{57.64}                                & \multicolumn{1}{l|}{48.03}                                & \multicolumn{1}{l|}{60.58}                                & \multicolumn{1}{l|}{53.09}                                & \multicolumn{1}{l|}{59.07}                                & \multicolumn{1}{l|}{71.59}                                & 73.00                                                     \\ \hline
\multicolumn{1}{|l|}{\textbf{Our TF Model}}               & \multicolumn{1}{l|}{97.67}                                & \multicolumn{1}{l|}{97.96}                                & \multicolumn{1}{l|}{27.63}                                     & \multicolumn{1}{l|}{33.33}                                     & \multicolumn{1}{l|}{43.08}                                & \multicolumn{1}{l|}{4974}                                & \multicolumn{1}{l|}{6349}                                & 66.32                                                     \\ \hline
\multicolumn{1}{|l|}{\textbf{Our RR Model}}                 & \multicolumn{1}{l|}{51.92}                                & \multicolumn{1}{l|}{52.75}                                & \multicolumn{1}{l|}{35.53}                                & \multicolumn{1}{l|}{32.43}                                & \multicolumn{1}{l|}{42.19}                                & \multicolumn{1}{l|}{40.17}                                & \multicolumn{1}{l|}{51.32}                                & 50.86                                                    \\ \hline 

\multicolumn{1}{|l|}{\textbf{Our TF Model (D)}}               & \multicolumn{1}{l|}{76.92}                                & \multicolumn{1}{l|}{64.29}                                & \multicolumn{1}{l|}{74.19}                                     & \multicolumn{1}{l|}{\textbf{84.38}}                                     & \multicolumn{1}{l|}{75.53}                                & \multicolumn{1}{l|}{72.97}                                & \multicolumn{1}{l|}{\textbf{95.40}}                                & \textbf{93.73}                                                     \\ \hline
\multicolumn{1}{|l|}{\textbf{Our RR Model (D)}}                 & \multicolumn{1}{l|}{90.87}                                & \multicolumn{1}{l|}{93.98}                                & \multicolumn{1}{l|}{\textbf{78.62}}                                & \multicolumn{1}{l|}{81.25}                                & \multicolumn{1}{l|}{\textbf{84.30}}                                & \multicolumn{1}{l|}{\textbf{87.15}}                                & \multicolumn{1}{l|}{91.51}                                & 92.86  
\\\hline


\end{tabular}
\caption{Comparison results after paraphrasing. Here D denotes the result after applying our proposed paraphrasing defence.}
\label{Tab: result_paraphrasing}
\end{table*}

\subsection{Results and Analysis}

Table \ref{Tab: result-1} shows the comparison results of the models when reviews are generated by GPT-4. It is evident from the results that our proposed TF and RR models outperform the other text detectors. In ICLR and NeurIPS dataset, our Token Frequency (TF) model surpasses the closest comparable model DEEP-FAKE with margins of 6.75 and 6.87 F1 points, RADAR by 29.45 and 26.28 F1 points, LLMDET by 29.69 and 30.64 F1 points. Whereas, Our Review Regeneration (RR) model outperforms DEEP-FAKE by 3.55 and 0.65 F1 points, RADAR by 26.25 and 20.06 F1 points, LLMDET by 26.49 and 24.42 F1 points and FAST DETECT by 8.76 and 15.03 F1 points


In the results reported above for the TF model, we considered tokens as adjectives, as this configuration yielded the best results. We also present the outcomes of the TF model when trained with tokens considered as adverbs or nouns in the Appendix Table \ref{Tab: TF_other}. Furthermore, we observe a similar distribution of results on reviews generated by GPT-3.5. We report the result in Appendix Table \ref{Tab: result-gpt3.5}.

\subsubsection{Effect of attacking AI-generated text detectors using Adjective Attack}
We report the results after performing adjective attack as described in Section \ref{Sec: AI word Attack} in Table \ref{Tab: result_att_adj}. It is evident from the table that the performance of each model dropped after the attack. In particular, for ICLR and NeurIPS respectively, the F1 score of RADAR dropped by 69.62\% and 68.18\%, LLMDET dropped by 6.46\% and 2.43\%, DEEP-FAKE dropped by 70.65\% and 88.10\%, and FAST DETECT dropped by 92.48\% and 98.29\%. Additionally, the F1 score of our TF model dropped by 79.88\% and 89.43\% for ICLR and NeurIPS, respectively, whereas for our RR model, it dropped by 25.56\% and 23.14\% for ICLR and NeurIPS, respectively.

The results reveal that this attack has significantly compromised the performance of our TF model, underscoring its vulnerability and limited resilience to such threats. The substantial decline in the F1-score can be attributed primarily to the model's reliance on token frequency patterns in AI-generated reviews. These patterns are effectively disrupted by synonym replacements leading to performance degradation. After the adjective attack, we observed that our RR model outperforms other AI text detectors, including our proposed TF model, achieving the highest F1 score of 71.81.





\subsubsection{Effect of attacking AI-generated text detectors using Paraphrasing Attack}
Next, we report the result after performing paraphrasing (See Appendix \ref{Appendix: Paraphrasing} for more details) on the AI-generated reviews. It is evident from the Table \ref{Tab: result_paraphrasing} that the result of each model dropped after the attack. In particular, for ICLR and NeurIPS, the F1 score of RADAR dropped by 7.10\% and 6.89\%, LLMDET dropped by 5.79\% and 3.62\%, DEEP-FAKE dropped by 18.19\% and 26.19\%, and FAST DETECT dropped by 39.69\% and 24.66\%. Additionally, F1 score of our TF model dropped by 56.92\% and 50.08\% for ICLR and NeurIPS respectively and RR model dropped by 56.41\% and 57.00\% for ICLR and NeurIPS respectively.

This effect on the TF model is not surprising, as it is based on AI token frequency and paraphrasing typically involves replacing words with their synonyms. For our RR model, we noted that paraphrasing caused both human-written and AI-written reviews to diverge further from the regenerated reviews. This increased dissimilarity could stem from various factors, including alterations in text structure, voice, tone, and vocabulary. If only human reviews had been paraphrased, we might have observed an improvement in performance due to a greater distinction between human-written and regenerated reviews. In our test set, which includes both AI-generated and human reviews, the similarity of AI-generated text decreased following paraphrasing, leading to a decline in overall performance.

\subsubsection{Results after Paraphrasing Defence}
Next, we report the result after performing paraphrasing Defence (See Section \ref{Section: Paraphrasing Defence} for more details) on both our proposed models on Table \ref{Tab: result_paraphrasing}. We observed improvements in both our TF and RR models. We also applied the defense to other AI text detection algorithms, observing no significant improvement or decrease in their results. These results are reported in Table \ref{Tab: result_defence}.  The performance of the TF model improved by 75.32\% for ICLR papers and 46.70\% for NeurIPS. Similarly, the performance of the RR model improved by 99.81\% for ICLR and 111.69\% for NeurIPS. 

These results indicate that our proposed RR model is more robust against different types of attacks and performs better than any other existing text detection algorithms.

\subsection{Human evaluation}
We also conducted human analyses to understand when and why our models fail.
Our model fails when paraphrasing alters the style or when AI-generated reviews closely resemble human writing, resulting in low similarity scores and incorrect predictions. We discuss this extensive error analysis in the Appendix \ref{Appendix: error}.

\section{Conclusion and Future Work}
In this work, we propose two methods to determine whether a review is written by a human or generated by AI. We found that our proposed TF model and the RR model outperform other AI text detectors under normal conditions. We stress test these detectors against token attack and paraphrasing. Furthermore, our proposed RR model is more robust and outperforms other methods. We then propose an effective defensive strategy to reduce the effect of paraphrasing on our models. Our findings suggest both of our proposed methods perform better than other AI text detectors. Also, while our proposed TF model performs better than the RR model without any attacks, our RR model is more robust against token attacks and paraphrasing attacks.


We hope that these findings will pave the way for more sophisticated and reliable AI detectors to prevent such misuse.
In future work, we aim to extend our analysis to other domains, such as Nuclear Physics, Medicine, and Social Sciences, and investigate domain-specific LLMs to enhance detection accuracy and explore the generalizability of our methods.

For further work, we aim to focus on cases where the reviewer writes parts of the review using AI.

\section*{Limitations} 

Our study primarily utilized GPT-4 and GPT-3.5 for generating AI texts, as GPT has been one of the most widely used LLMs for long-context content generation. We recommend that future practitioners choose the LLM that best aligns with the language model likely used to generate their target corpus, to accurately reflect usage patterns at the time of its creation. Our methods are specifically designed for reviews completely written by AI. It is possible, however, that a reviewer may outline several bullet points related to a paper and use ChatGPT to expand these into full paragraphs. We suggest exploring this aspect in future research.

\section*{Ethics Statement}

We have utilized the open source dataset for this study. We do not claim that the use of AI tools for review papers is necessarily bad or good, nor do we provide definitive proof that reviewers are employing ChatGPT to draft reviews. The primary purpose of this system is to assist editors by identifying potentially AI-generated reviews, and is intended only for editors' internal usage, not for authors or reviewers.

Our RR model requires regenerated review to be generated from paper using LLM. Also, open-sourced LLMs running locally will not have any concerns. OpenAI implemented a Zero Data Retention policy to ensure the security and privacy of data. Additionally, users can control the duration of data retention through ChatGPT Enterprise\footnote{\url{https://openai.com/index/introducing-chatgpt-enterprise/}}. Also, nowadays, many papers are submitted to arXiv and are publicly available\footnote{\url{https://arxiv.org/}}. However, editors and chairs should use this tool with caution, considering the potential risks to privacy and anonymity.

The system cannot detect all AI-generated reviews and may produce false negatives, so editors should not rely on it exclusively. It is meant to assist, but results must be verified and analyzed carefully before making any decisions. We hope that our data and analyses will facilitate constructive discussions within the community and help prevent the misuse of AI.

\section*{Acknowledgement}
Sandeep Kumar acknowledges the Prime Minister Research Fellowship (PMRF) program of the Govt of India for its support. 

\bibliography{custom}
\bibliographystyle{acl_natbib}

\appendix

\section{Dataset} \label{Appendix: dataset}

We generated a fake review for each paper using both GPT-3.5 and GPT-4. We gave the prompt template similar to both of the conference style of reviews. We also generated regenerated reviews for this task.

We discuss the dataset in more details in the Appendix Section

Below is the prompt we used for generating AI-generated review ICLR 2022 reviews:
\begin{tcolorbox}[colback=gray!10!white, colframe=gray!60!black]
    \textbf{System}: You are a research scientist reviewing a scientific paper. \\
    \textbf{User}: Read the following paper and write a thorough peer-review in the following format:  \\
        1) Summary of the paper \\
        2) Main review \\
        3) Summary of the review
    \begin{quote}
        [paper text]
    \end{quote}
\end{tcolorbox}

Below is the prompt we used for generating AI-generated review NeurIPS 2022 reviews:

\begin{tcolorbox}[colback=gray!10!white, colframe=gray!60!black]
    \textbf{System}: You are a research scientist reviewing a scientific paper. \\
    \textbf{User}: Read the following paper and write a thorough peer-review in the following format:  \\
        1) Summary (avg word length 100) \\
        2) Strengths and weaknesses \\
        3) Questions \\
        4) Limitations (in short)
    \begin{quote}
        [paper text]
    \end{quote}
\end{tcolorbox}

Below is the prompt we used for generating AI-regenerated review ICLR 2022 reviews:-
\begin{tcolorbox}[colback=gray!10!white, colframe=gray!60!black]
    \textbf{System}: You are a research scientist reviewing a scientific paper. \\
    \textbf{User}: Your task is to draft a high-quality peer-review in the below format:  \\
        1) Summarize the paper. \\
        2) List strong and weak points of the paper, Question and Feedback to the author. Be as comprehensive as possible.\\
        3) Write review summary (Provide supporting arguments for your recommendation).
    \begin{quote}
        [paper text]
    \end{quote}
\end{tcolorbox}

To generate AI-regenerated reviews, we used prompts that were very distinct from those we used to generate AI reviews for training. The reason for this approach is that a reviewer may write any kind of prompt, which could be very different from the prompts we used for training.

Below is the prompt we used for generating AI regenerated review  NeurIPS 2022 reviews :-
\begin{tcolorbox}[colback=gray!10!white, colframe=gray!60!black]
    \textbf{System}: You are a research scientist reviewing a scientific paper. \\
    \textbf{User}: Your task is to draft a high-quality peer-review in the below format:  \\
        1) Briefly summarize the paper and its contributions \\
        2) Please provide a thorough assessment of the strengths and weaknesses of the paper \\
        3) Please list up and carefully describe any questions and suggestions for the authors
        4) Limitations: Have the authors adequately addressed the limitations and potential negative societal impact of their work? If not, please include constructive suggestions for improvement. Write in few lines only
    \begin{quote}
        [paper text]
    \end{quote}
\end{tcolorbox}
\section{Error Analysis} \label{Appendix: error}
We conducted an analysis of the predictions made by our proposed baseline to identify the areas where it most frequently fails.

\subsection{Challenges after paraphrasing:} 
Our regeneration-based approach sometimes fails when it processes a paraphrased review. Paraphrasing can alter the semantics of a review to some extent, leading to discrepancies with our reverse-generated reviews. Consequently, our model may incorrectly predict these as human-written rather than AI-generated. Our proposed defense strategy corrects only the tokens that have been changed during paraphrasing. However, when the paraphrasing significantly alters the style, our RR model fails. 

\subsection{Sometimes Regenerated review and AI written reviews are similar:}
Our RR model works on the similarity of review and Regenerated review. We found the model fails when LLM generates a review that is very much similar to human writing. In those cases, we found that the similarity score tends to be low, leading to the model's failure. This suggests the model may struggle to differentiate human-like AI-generated text.



\section{Token Attack} \label{Appendix: token_attack}
Below is an example of how impactful various attacks can be when replacing words in a review:-
After reviewing all the attacks, we observe that the adjective attack produced more logical changes compared to the others. For example, in the noun attack, `model' was replaced with 'pose,' 'learning' with 'discovery,' 'performance' with 'execution,' and 'datasets' with 'information sets,' which are not very meaningful and thus make the attack less effective. Replacing words can cause significant changes in the meaning of a review and can even alter the context. So we used only the adjective attack for our experiments.

\begin{tcolorbox}[colback=gray!10!white, left=0mm, 
                 right=0mm,  colframe=gray!60!black]
    \textbf{Actual Sentence: } The \textcolor{purple}{model} is evaluated in both reinforcement \textcolor{purple}{learning} and vision settings, \textcolor{blue}{showcasing} \textcolor{red}{significant} \textcolor{purple}{performance} boosts in \textcolor{blue}{tasks} such \textcolor{blue}{as} DMC Suite with distractors and CIFAR-10/STL10 \textcolor{blue}{datasets}. \\

    \textbf{Adjective: } The model is evaluated in both reinforcement learning and vision settings, showcasing \textcolor{red}{substantial} performance boosts in tasks such as DMC Suite with distractors and CIFAR-10/STL-10 datasets. \\

    \textbf{Noun: } The \textcolor{purple}{pose} is evaluated in both reinforcement \textcolor{purple}{discover} and vision scene, showcasing significant \textcolor{purple}{execution} boosts in project such as DMC Suite with distractors and CIFAR-10/STL-10 \textcolor{purple}{informationsets}. \\

    \textbf{Adverb: } The model is evaluated in both reinforcement learning and vision settings, \textcolor{blue}{showcequallying} significant performance boosts in \textcolor{blue}{tequallyks} such \textcolor{blue}{equally} DMC Suite with distractors and CIFAR-10/STL-10 \textcolor{blue}{datequallhowevers}
\end{tcolorbox}


\section{Baseline Comparison} \label{Appendix: baseline}
\subsection{\textbf{RADAR \cite{DBLP:conf/nips/HuCH23}}}
 The way RADAR works is as follows - First, an AI-text corpus is generated from a target (frozen) language model from a human-text corpus. The next step is followed by introduction of a paraphraser (a tunable language model) and a detector (a separate tunable language model).
 In the training stage, the detector's objective is to distinguish between human-generated text and AI-generated text, whereas the paraphraser's goal is to rephrase AI-generated text to avoid detection. The model parameters of the paraphraser and detector are updated in an adversarial learning manner. During the evaluation (testing) phase, the deployed detector utilizes its training to assess the probability of content being AI-generated for any given input instance.

 \subsection{\textbf{LLMDET \cite{wu2023llmdet}}:}
 The overall framework of the system consists of two main components - 1) Dictionary creation and 2) Text detection. The main idea was to make use of the perplexity as a measurement of identifying
the generated text from different LLMs. 
    So the dictionary had \textit{n}-grams as keys and the next token probablities as values. The dictionary serves as prior information during the detection process. Since the dictionary of n-grams and their probabilities was obtained, it enabled the utilization of the corresponding dictionary of each model as prior information for third-party detection, facilitating the calculation of the proxy perplexity of the text being detected on each model. Proxy perplexity was then used as a feature into a trained text classifier, the corresponding detection results were obtained.

\subsection{\textbf{DEEP-FAKE \cite{li2023deepfake}}}
To determine whether machine-generated text can be discerned from human-written content, the collected data was categorized into six settings based on the sources used for model training and evaluation. These settings progressively increased the difficulty of detection. The classifier then assigned a probability to each text, indicating the likelihood of it being authored by humans or generated by language model models (LLMs).
AvgRec (average recall) was the principal metric, calculated as the average score between the recall on human-written texts (HumanRec) and the recall on machine-generated texts (MachineRec).

\subsection{\textbf{FAST-DETECT GPT \cite{bao2023fast}}}
The model comprises of a three-fold architecture - 1) Revealing and confirming a novel conjecture that humans and machines exhibit distinct word selection patterns within a given context. 2)  Employing conditional probability curvature as a fresh characteristic to identify machine-generated text, thereby reducing the detection expenses by a factor of 2 orders of magnitude. 3) Attaining the highest average detection accuracy in both white-box and black-box environments and comparing to current zero-shot text detection systems.

\section{Paraphrasing} \label{Appendix: Paraphrasing}

We performed paraphrasing by providing prompts to the Gemini model \cite{team2023gemini}. We have provided example of paraphrased review in \hyperref[tab: paraphrased_eg]{table 6}

We used the following prompt for generating paraphrased text:-
\begin{tcolorbox}[colback=gray!10!white, colframe=gray!60!black]
    \textbf{System}: You are a paraphraser. \\
    \textbf{User}: Paraphrase the following review: 
        
    \begin{quote}
        [Review]
    \end{quote}
\end{tcolorbox}

\begin{table*}

\begin{center}
\begin{tabular}{|p{15.5cm}|}
\hline
    \textit{\textbf{Actual Review}}
 \\
\hline
1. \textcolor{purple}{Summary of the Paper:-} The paper explores the incorporation of higher-order dynamics specifically second derivatives into neural models to improve the estimation of cardiac pulse dynamics. The focus is on video-based vital sign measurement particularly Photoplethysmography (PPG) using deep learning architectures. The research demonstrates that optimizing for second derivatives in the loss function enhances the estimation of waveform morphology crucial for clinically significant scenarios such as left ventricle ejection time (LVET) intervals. The study uses simulationgenerated data for training due to the scarcity of labeled real data and evaluates model performance against a real dataset.  

2.\textcolor{purple}{Main Review:-} The paper presents a novel approach by considering higher-order dynamics in the context of video-based cardiac measurements a crucial step towards capturing subtle variations in arterial health indicators. The methodology is wellstructured building on existing literature and providing a clear rationale for exploring multiderivative learning objectives in neural models. The experiments are detailed utilizing synthetic data for training and real data for evaluation demonstrating the feasibility and effectiveness of incorporating second derivatives in deep learning models for cardiac pulse estimation. The systematic evaluation conducted by the authors is rigorous focusing on quantitative metrics such as heart rate (HR) and left ventricle ejection time (LVET) intervals to compare different model configurations. The results indicate improvements in LVET estimation when secondderivative frames are included highlighting the benefits of considering higher-order dynamics in the estimation process. The study also includes a qualitative analysis providing visual representations of the predicted signals and comparing them to ground truth data enhancing the interpretation of the quantitative results. Overall the inclusion of both quantitative and qualitative assessments strengthens the papers conclusions and provides a comprehensive understanding of the proposed methodology. 

3. \textcolor{purple}{Summary of the Review:-} The paper makes a valuable contribution to the field of video-based cardiac measurement by emphasizing the importance of higher-order dynamics in accurately estimating vital sign parameters. The research framework is wellbuilt supported by a thorough literature review detailed experiments and insightful conclusions. The results demonstrate the significance of optimizing for second derivatives in neural models improving the estimation of clinically relevant waveform morphology especially in scenarios like LVET intervals. Overall the paper presents a compelling case for incorporating multiderivative learning objectives to enhance the accuracy of video-based vital sign measurements.
\\

\hline
\textit{\textbf{Paraphrased Review}} \\
\hline
1. \textcolor{purple}{Summary of the Paper:}

The paper investigates including second derivatives (higher-order dynamics) in neural models to accurately measure cardiac pulse signals using video (Photoplethysmography or PPG). By focusing on these higher-order dynamics, the models can better capture subtle changes in these signals, leading to improved estimates of waveform characteristics. Since real-world data is limited, the models are trained on simulated data and then tested on real data.

2. 
\textcolor{purple}{Main Review:}

This novel approach is a step forward in understanding subtle fluctuations in cardiac health indicators through video-based measurements. The paper builds on existing research, providing a strong rationale for incorporating higher-order dynamics into neural models for cardiac pulse estimation. The experiments use both simulated and real data, demonstrating the effectiveness of including second derivatives in these models.

Rigorous evaluation using metrics like heart rate and LVET intervals shows that incorporating second derivatives improves LVET estimation. Qualitative analysis, comparing predicted signals to ground truth, further supports these findings.

 3. \textcolor{purple}{Summary of the Review:}

This paper significantly contributes to video-based cardiac measurements by highlighting the importance of higher-order dynamics in accurately estimating vital sign parameters. The well-structured research framework, detailed experiments, and insightful conclusions demonstrate the value of optimizing for second derivatives in neural models. This approach enhances waveform morphology estimation, especially for clinically important measures like LVET intervals, making it a valuable addition to the field.
\\
\hline
\end{tabular}
\label{tab: paraphrased_eg}
\end{center}
\caption{Examples of Actual and Paraphrased Review.}
\end{table*}

\section{Proof Reading} \label{Appendix: ProofReading}
We randomly picked up 100 human reviews from our test set and proofread them using "gpt-4-turbo" model. We gave two different prompts to the model: 

\begin{tcolorbox}[colback=gray!10!white, colframe=gray!60!black]
    \textbf{Prompt-1}: You have to proof-read the provided review, don't write anything additional except the review in that same format, but just proof-read it: 
        
    \begin{quote}
        [Review]
    \end{quote}
    \textbf{Prompt-2}: Modify the review to make it more clear and coherent. Ensure that there are no grammatical or spelling errors: 
        
    \begin{quote}
        [Review]
    \end{quote} 
\end{tcolorbox}

We found no False Positive by either our RR model or our proposed TF model in our first prompt, and no False Positive by our RR model and 6 False Positive by TF model in our second prompt, which shows both models have very little effect on proofreading.

\begin{table*}[]
\begin{tabular}{|l|llllllll|}
\hline
\multirow{2}{*}{\textbf{Model}}
                                & \multicolumn{2}{c|}{\textbf{Precision}}                                 & \multicolumn{2}{c|}{\textbf{Recall}}                                    & \multicolumn{2}{c|}{\textbf{F1 - Score}}                                & \multicolumn{2}{c|}{\textbf{Accuracy}}             \\ \cline{2-9} 
                                & \multicolumn{1}{l|}{\textbf{ICLR}} & \multicolumn{1}{l|}{\textbf{NeurIPS}} & \multicolumn{1}{l|}{\textbf{ICLR}} & \multicolumn{1}{l|}{\textbf{NeurIPS}} & \multicolumn{1}{l|}{\textbf{ICLR}} & \multicolumn{1}{l|}{\textbf{NeurIPS}} & \multicolumn{1}{l|}{\textbf{ICLR}} & \textbf{NeurIPS} \\ \hline
\textbf{RADAR}                  & \multicolumn{1}{l|}{29.58}        & \multicolumn{1}{l|}{31.75}        & \multicolumn{1}{l|}{79.60}        & \multicolumn{1}{l|}{93.05}        & \multicolumn{1}{l|}{69.29}        & \multicolumn{1}{l|}{70.72}        & \multicolumn{1}{l|}{60.12}        & 62.37        \\ \hline
\textbf{LLMDET}                 & \multicolumn{1}{l|}{19.38}        & \multicolumn{1}{l|}{18.64}        & \multicolumn{1}{l|}{\textbf{98.03}}        & \multicolumn{1}{l|}{\textbf{98.61}}        & \multicolumn{1}{l|}{32.36}        & \multicolumn{1}{l|}{31.35}        & \multicolumn{1}{l|}{22.13}        & 21.46        \\ \hline
\textbf{DEEP-FAKE}              & \multicolumn{1}{l|}{76.68}        & \multicolumn{1}{l|}{75.81}        & \multicolumn{1}{l|}{97.37}        & \multicolumn{1}{l|}{0.9792}        & \multicolumn{1}{c|}{85.80}        & \multicolumn{1}{l|}{85.45}        & \multicolumn{1}{l|}{86.35}        & 86.32        \\ \hline
\textbf{FAST DETECT}               & \multicolumn{1}{l|}{84.88}        & \multicolumn{1}{l|}{82.31}        & \multicolumn{1}{l|}{96.05}        & \multicolumn{1}{l|}{84.03}        & \multicolumn{1}{l|}{90.12}        & \multicolumn{1}{l|}{83.16}        & \multicolumn{1}{l|}{96.00}        & 93.81        \\ \hline
\multicolumn{1}{|l|}{\textbf{Our RR Model}} & \multicolumn{1}{l|}{\textbf{99.34}}                                & \multicolumn{1}{l|}{\textbf{95.14}}                                & \multicolumn{1}{l|}{93.79}                                & \multicolumn{1}{l|}{92.57}                                & \multicolumn{1}{l|}{\textbf{96.49}}                                & \multicolumn{1}{l|}{\textbf{93.84}}                                & \multicolumn{1}{l|}{\textbf{98.49}}                                & \textbf{97.36} 
\\ \hline
\end{tabular}
\caption{Comparison Result of proposed Review Regeneration technique; Here the AI-generated reviews and regenerated reviews are generated by GPT-3.5. ; RR: Review Regeneration; TF: Token Frequency.}
\label{Tab: result-gpt3.5}
\end{table*}

\begin{table*}[]
\begin{tabular}{|l|llllllll|}
\hline
                              & \multicolumn{2}{c|}{\textbf{Precision}}                                                         & \multicolumn{2}{c|}{\textbf{Recall}}                                                            & \multicolumn{2}{c|}{\textbf{F1-Score}}                                                          & \multicolumn{2}{c|}{\textbf{Accuracy}}                                                          \\ \cline{2-9} 
\multirow{-3}{*}{\textbf{Model}} & \multicolumn{1}{c|}{\textbf{ICLR}} & \multicolumn{1}{c|}{\textbf{NeurIPS}} & \multicolumn{1}{c|}{\textbf{ICLR}} & \multicolumn{1}{c|}{\textbf{NeurIPS}} & \multicolumn{1}{c|}{\textbf{ICLR}} & \multicolumn{1}{c|}{\textbf{NeurIPS}} & \multicolumn{1}{c|}{\textbf{ICLR}} & \multicolumn{1}{c|}{\textbf{NeurIPS}} \\ \hline
\textbf{ADJECTIVE}                                       & \multicolumn{1}{l|}{\textbf{99.99}}             & \multicolumn{1}{l|}{\textbf{99.99}}             & \multicolumn{1}{l|}{99.80}         & \multicolumn{1}{l|}{99.30}         & \multicolumn{1}{l|}{99.99}         & \multicolumn{1}{l|}{99.65}        & \multicolumn{1}{l|}{\textbf{99.92}}        & 99.82                                                     \\ \hline
\textbf{NOUN}                                            & \multicolumn{1}{l|}{91.45}        & \multicolumn{1}{l|}{\textbf{99.99}}             & \multicolumn{1}{l|}{\textbf{99.99}}             & \multicolumn{1}{l|}{\textbf{99.99}}             & \multicolumn{1}{l|}{95.53}        & \multicolumn{1}{l|}{\textbf{99.99}}             & \multicolumn{1}{l|}{98.50}         & \textbf{99.99}                                                          \\ \hline
\textbf{ADVERB}                                          & \multicolumn{1}{l|}{93.42}                                & \multicolumn{1}{l|}{90.97}                                & \multicolumn{1}{l|}{89.86}                                & \multicolumn{1}{l|}{90.35}                                & \multicolumn{1}{l|}{91.61}                                & \multicolumn{1}{l|}{90.66}                                & \multicolumn{1}{l|}{97.00}                                  & 95.16                                                     \\ \hline
\end{tabular}
\caption{Result of Token Frequency based Approach. Here the fake review is generated by prompting GPT-4.}
\label{Tab: TF_other}
\end{table*}

\begin{table*}[]
\begin{tabular}{|l|llllllll|}
\hline
                              & \multicolumn{2}{c|}{\textbf{Precision}}                                                         & \multicolumn{2}{c|}{\textbf{Recall}}                                                            & \multicolumn{2}{c|}{\textbf{F1-Score}}                                                          & \multicolumn{2}{c|}{\textbf{Accuracy}}                                                          \\ \cline{2-9} 
\multirow{-3}{*}{\textbf{Model}} & \multicolumn{1}{c|}{\textbf{ICLR}} & \multicolumn{1}{c|}{\textbf{NeurIPS}} & \multicolumn{1}{c|}{\textbf{ICLR}} & \multicolumn{1}{c|}{\textbf{NeurIPS}} & \multicolumn{1}{c|}{\textbf{ICLR}} & \multicolumn{1}{c|}{\textbf{NeurIPS}} & \multicolumn{1}{c|}{\textbf{ICLR}} & \multicolumn{1}{c|}{\textbf{NeurIPS}} \\ \hline
\textbf{ADJECTIVE}                                       & \multicolumn{1}{l|}{\textbf{99.99}}             & \multicolumn{1}{l|}{\textbf{99.99}}             & \multicolumn{1}{l|}{98.70}         & \multicolumn{1}{l|}{\textbf{99.32}}         & \multicolumn{1}{l|}{\textbf{99.35}}         & \multicolumn{1}{l|}{\textbf{99.66}}        & \multicolumn{1}{l|}{\textbf{99.77}}        & \textbf{99.82}                                                     \\ \hline
\textbf{NOUN}                                            & \multicolumn{1}{l|}{98.69}        & \multicolumn{1}{l|}{99.99}             & \multicolumn{1}{l|}{\textbf{99.34}}             & \multicolumn{1}{l|}{97.92}             & \multicolumn{1}{l|}{99.02	}        & \multicolumn{1}{l|}{98.95}             & \multicolumn{1}{l|}{99.65}         & 99.46                                                         \\ \hline
\textbf{ADVERB}                                          & \multicolumn{1}{l|}{96.55}                                & \multicolumn{1}{l|}{97.24}                                & \multicolumn{1}{l|}{92.11}                                & \multicolumn{1}{l|}{97.92}                                & \multicolumn{1}{l|}{94.28}                                & \multicolumn{1}{l|}{97.58}                                & \multicolumn{1}{l|}{98.03}                                  & 98.75                                                     \\ \hline
\end{tabular}
\caption{Result of Token Frequency-based Approach. Here the fake review is generated by prompting GPT-3.5.}
\label{Tab: TF_other}
\end{table*}

\begin{table*}[ht]
\begin{tabular}{|l|llllllll|}
\hline

                                 & \multicolumn{2}{c|}{\textbf{Precision}}                                                         & \multicolumn{2}{c|}{\textbf{Recall}}                                                            & \multicolumn{2}{c|}{\textbf{F1-Score}}                                                          & \multicolumn{2}{c|}{\textbf{Accuracy}}                                                          \\ \cline{2-9} 

\multirow{-3}{*}{\textbf{Model}} & \multicolumn{1}{c|}{\textbf{ICLR}} & \multicolumn{1}{c|}{\textbf{NeurIPS}} & \multicolumn{1}{c|}{\textbf{ICLR}} & \multicolumn{1}{c|}{\textbf{NeurIPS}} & \multicolumn{1}{c|}{\textbf{ICLR}} & \multicolumn{1}{c|}{\textbf{NeurIPS}} & \multicolumn{1}{c|}{\textbf{ICLR}} & \multicolumn{1}{c|}{\textbf{NeurIPS}}                                                   \\ \hline
\textbf{RADAR}                                           & \multicolumn{1}{l|}{14.47}        & \multicolumn{1}{l|}{10.42}        & \multicolumn{1}{l|}{59.46}        & \multicolumn{1}{l|}{57.69}        & \multicolumn{1}{l|}{23.28}        & \multicolumn{1}{l|}{17.65}        & \multicolumn{1}{l|}{52.30}        & 51.39                                                    \\ \hline

\textbf{LLMDET}                                          & \multicolumn{1}{l|}{97.37}        & \multicolumn{1}{l|}{95.77}        & \multicolumn{1}{l|}{50.68}        & \multicolumn{1}{l|}{49.64}        & \multicolumn{1}{l|}{66.67}        & \multicolumn{1}{l|}{65.38}        & \multicolumn{1}{l|}{51.32}        & 50.00                                                    \\ \hline
\textbf{DEEP-FAKE}                                       & \multicolumn{1}{l|}{35.38}        & \multicolumn{1}{l|}{44.44}                                & \multicolumn{1}{l|}{71.88}        & \multicolumn{1}{l|}{59.26}                                & \multicolumn{1}{l|}{47.42}        & \multicolumn{1}{l|}{50.79}                                & \multicolumn{1}{l|}{55.91}        & 56.94                                                     \\ \hline
\textbf{FAST DETECT}                                     & \multicolumn{1}{l|}{5.26}                                & \multicolumn{1}{l|}{7.64}                                & \multicolumn{1}{l|}{80.00}                                & \multicolumn{1}{l|}{84.62}                                & \multicolumn{1}{l|}{9.88}                                & \multicolumn{1}{l|}{14.01}                                & \multicolumn{1}{l|}{67.84}                                & 68.31                                                     \\ \hline

\multicolumn{1}{|l|}{\textbf{Our TF Model}}               & \multicolumn{1}{l|}{76.92}                                & \multicolumn{1}{l|}{64.29}                                & \multicolumn{1}{l|}{74.19}                                     & \multicolumn{1}{l|}{84.38}                                     & \multicolumn{1}{l|}{75.53}                                & \multicolumn{1}{l|}{72.97}                                & \multicolumn{1}{l|}{95.40}                                & 93.73                                                     \\ \hline
\multicolumn{1}{|l|}{\textbf{Our RR Model}}                 & \multicolumn{1}{l|}{90.87}                                & \multicolumn{1}{l|}{93.98}                                & \multicolumn{1}{l|}{78.62}                                & \multicolumn{1}{l|}{81.25}                                & \multicolumn{1}{l|}{84.30}                                & \multicolumn{1}{l|}{87.15}                                & \multicolumn{1}{l|}{91.51}                                & 92.86  
                                                  \\ \hline


\end{tabular}
\caption{Comparison results after paraphrasing applying Paraphrasing defence.}
\label{Tab: result_defence}
\end{table*}

\end{document}